\documentclass[runningheads]{llncs}

\usepackage[T1]{fontenc}
\usepackage{balance}
\usepackage{float}
\usepackage{graphicx}
\usepackage{amsmath}
\usepackage{amssymb}
\usepackage{booktabs}
\usepackage{subcaption}
\usepackage{mathpazo}
\begin{document}

\title{WaveFormer: Wavelet Embedding Transformer \\ for Biomedical Signals}

\author{Habib Irani \and
Bikram De \and
Vangelis Metsis}
%


\authorrunning{H. Irani et al.}
%
\institute{Texas State University, San Marcos, TX 78666, USA 
\email{\{habibirani,bikramkumarde,vmetsis\}@txstate.edu}}
%
%

\maketitle

\begin{abstract}


 Biomedical signal classification presents unique challenges due to long sequences, complex temporal dynamics, and multi-scale frequency patterns that are poorly captured by standard transformer architectures. We propose WaveFormer, a transformer architecture that integrates wavelet decomposition at two critical stages: embedding construction, where multi-channel Discrete Wavelet Transform (DWT) extracts frequency features to create tokens containing both time-domain and frequency-domain information, and positional encoding, where Dynamic Wavelet Positional Encoding (DyWPE) adapts position embeddings to signal-specific temporal structure through mono-channel DWT analysis. We evaluate WaveFormer on eight diverse datasets spanning human activity recognition and brain signal analysis, with sequence lengths ranging from 50 to 3000 timesteps and channel counts from 1 to 144. Experimental results demonstrate that WaveFormer achieves competitive performance through comprehensive frequency-aware processing. Our approach provides a principled framework for incorporating frequency-domain knowledge into transformer-based time series classification.


\end{abstract}

\section{Introduction}

Time series classification remains a fundamental problem in machine learning with applications spanning healthcare diagnostics, industrial monitoring, financial forecasting, and human activity recognition. Among these domains, biomedical signal analysis presents particularly challenging characteristics: signals are multivariate with complex inter-channel dependencies, exhibit non-stationary dynamics where statistical properties evolve over time, and contain discriminative information distributed across multiple temporal and frequency scales. Traditional machine learning approaches, while effective for specific tasks, often require extensive domain expertise for feature engineering and struggle to capture the hierarchical temporal structure inherent in physiological data.

The Transformer architecture~\cite{vaswani2017attention} has emerged as a powerful paradigm for sequence modeling, primarily due to its self-attention mechanism that enables direct modeling of long-range dependencies without the recurrence bottleneck of RNNs or the limited receptive field of CNNs. However, standard Transformers face two critical limitations when applied to time series. First, the self-attention mechanism is inherently permutation invariant, requiring explicit positional information to capture temporal ordering.

Second, conventional input embedding strategies apply linear projections to raw time-domain values, implicitly assuming that subsequent attention layers will automatically discover relevant features. This assumption is questionable for complex signals where clinically or physically meaningful patterns manifest in specific frequency bands. For instance, brain activity measured via EEG exhibits well-characterized rhythms (alpha, beta, gamma bands) that are not explicitly represented in raw amplitude sequences. Without explicit multi-scale feature extraction, models must implicitly learn frequency decomposition, which is computationally expensive and may converge to suboptimal representations. 

\begin{figure}
\centering
\includegraphics[width=0.75\columnwidth]{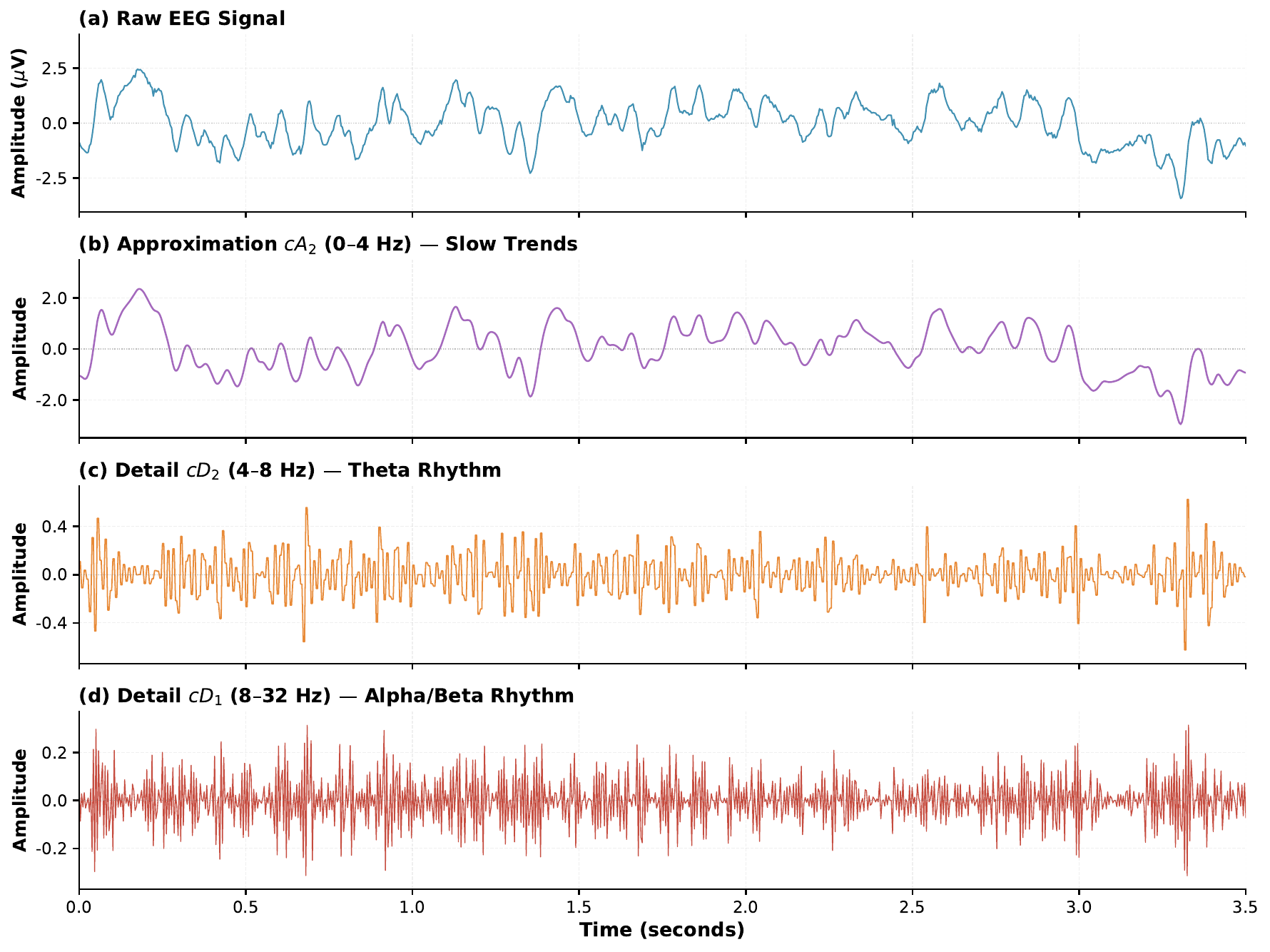}
\caption{Example of multi-scale wavelet decomposition of an EEG signal (SelfRegulationSCP1). Different frequency bands capture distinct physiological patterns: approximation coefficients encode slow trends while detail coefficients isolate specific neural rhythms.}
\label{fig:wavelet_decomposition1}
\end{figure}

As illustrated in Figure~\ref{fig:wavelet_decomposition1}, biomedical signals contain discriminative information across multiple frequency scales, with different bands corresponding to distinct physiological patterns. 
This multi-scale structure motivates an architecture that explicitly preserves and exploits frequency-band information rather than relying on raw amplitudes alone.

Recent advances have begun addressing these challenges through various approaches. 
Convolutional Transformers~\cite{foumani2023series2vec} combine CNN feature 
extraction with transformer attention, while patch-based methods~\cite{nie2023time} 
adapt Vision Transformer techniques to time series.

We introduce \textbf{WaveFormer}, a dual-stage wavelet-enhanced transformer architecture that employs the Discrete Wavelet Transform (DWT) as a principled mathematical framework for explicit multi-scale feature extraction. The wavelet transform naturally decomposes signals into multi-scale frequency components, providing approximation coefficients that capture coarse temporal trends and detail coefficients that capture fine-scale variations. By integrating wavelet analysis at both stages of the embedding pipeline, WaveFormer ensures consistency in multi-scale representation: tokens are enriched with explicit frequency-domain features while simultaneously receiving positional encodings creating a rich representation that dynamically adapts to local behavior.

The primary contributions of this work are:
\begin{itemize}
    \item A novel dual-stage wavelet integration architecture that coherently 
    addresses both feature extraction and positional encoding through multi-scale 
    signal decomposition.
    \item A wavelet-enhanced patch embedding that decomposes temporal information into approximation (low-frequency) and detail (high-frequency) coefficients that are fused with raw time-domain patches for dual-path feature representation.
    \item Comprehensive empirical validation on eight datasets demonstrating 
    consistent improvements over state-of-the-art baselines and analysis revealing that performance gains scale with sequence length, supporting the theoretical motivation for 
    explicit multi-scale decomposition.
\end{itemize}

\section{Literature Review}

Early deep learning approaches to time series relied on Recurrent Neural Networks ~\cite{hochreiter1997long}, which naturally model sequential dependencies through hidden state evolution. While LSTM and GRU architectures addressed the vanishing gradient problem, they remain fundamentally limited by sequential processing requirements and struggle with very long-range dependencies. Convolutional approaches, particularly InceptionTime ~\cite{fawaz2020inceptiontime}, demonstrated that hierarchical feature extraction through multi-scale convolutional filters could achieve competitive or superior performance with significantly improved computational efficiency. These architectures leverage strong inductive biases, local pattern recognition, and translation invariance but lack mechanisms for capturing global, long-range dependencies without increasing network depth substantially. These architectures leverage strong inductive biases for local pattern recognition and translation invariance, but struggle with global, long-range dependencies due to their limited receptive fields. Despite techniques like dilated convolutions \cite{bai2018tcn} that expand receptive fields, CNNs require substantial depth to capture dependencies across entire sequences. Recent work has addressed this limitation by incorporating broader temporal context to improve prediction stability \cite{irani2024enhancing}, though fundamental scalability challenges remain for very long sequences.

The introduction of Transformers to time series ~\cite{zerveas2021transformer} promised to overcome both the sequential bottleneck of RNNs and the limited receptive field of CNNs through global self-attention. However, the quadratic complexity of attention with respect to sequence length necessitated architectural innovations. 
PatchTST ~\cite{cordonnier2021differentiable} addressed scalability by treating contiguous subsequences as tokens, reducing sequence length while maintaining temporal resolution. Other notable architectures include Informer~\cite{zhou2021informer} with its ProbSparse attention mechanism, and Autoformer~\cite{wu2021autoformer}, which explicitly incorporates time series decomposition. These models primarily focus on architectural modifications to the attention mechanism rather than on optimizing the input representation pipeline.

\subsection{Positional Encoding: From Fixed to Signal-Aware Approach}

Positional encoding is fundamental to transformers due to their permutation-invariant attention mechanism. Conventional approaches employ either fixed sinusoidal patterns or learned embeddings~\cite{vaswani2017attention,su2024roformer}, both of which assign identical positional information to sequence indices regardless of signal content. A recent comprehensive survey~\cite{irani2025positional} systematically categorized positional encoding methods across transformer applications.

Relative positional encoding~\cite{shaw2018self} advanced the field by recognizing that pairwise temporal relationships often matter more than absolute positions, with methods like T5~\cite{raffel2020exploring} introducing bucketing strategies for parameter efficiency. However, these approaches still compute positional biases purely from index differences, ignoring signal characteristics. Recent work has begun exploring signal-aware schemes that derive positional information from content itself. Dynamic Wavelet Positional Encoding (DyWPE)~\cite{irani2025dywpe} represents a significant advance in this direction, employing wavelet decomposition with dynamic gating to generate position embeddings that reflect local frequency content, achieving consistent improvements, particularly on long sequences and biomedical signals. 

\subsection{Wavelet-Based Architectures for Time Series}

Wavelet transforms have long been recognized in signal processing for their ability to decompose signals into multi-scale frequency components, making them particularly suitable for analyzing non-stationary time series. The integration of wavelets into deep learning and embedding is comprehensively reviewed in~\cite{irani2025time}. Recent deep learning architectures have integrated wavelets in various ways to enhance multi-resolution feature learning.

Subsequent work introduced learnable wavelet parameters to enable end-to-end optimization. Wang et al.~\cite{wang2018mwdn} proposed the Multilevel Wavelet Decomposition Network (mWDN), which makes wavelet decomposition fully trainable within neural networks. More recently, Feng et al.~\cite{feng2025mwdn} extended this for time series forecasting with MWDN, which employs iterative wavelet decomposition with dual-branch architectures that separately model seasonal (high-frequency) and trend (low-frequency) components. 

For vision transformers, Yao et al.~\cite{yao2022wavevit} introduced Wave-ViT, which applies DWT for lossless downsampling of keys and values in self-attention. By decomposing into low-frequency approximations and high-frequency details, Wave-ViT preserves texture information typically lost in pooling-based downsampling, achieving 85.5\% ImageNet accuracy while maintaining computational efficiency. The architecture leverages inverse DWT to reconstruct high-resolution features with enlarged receptive fields, though wavelets are applied only to the attention mechanism rather than to positional encoding or input features.

Despite these advances, existing wavelet-based architectures employ wavelets in isolation, either for feature extraction, attention downsampling, or preprocessing, without systematically integrating multi-scale analysis across all representational stages.

\section{Methodology}

\subsection{Problem Formulation and Architecture Overview}

Given a multivariate time series $\mathbf{X} \in \mathbb{R}^{L \times C}$ with sequence length $L$ and $C$ channels, our objective is to learn a classification function $f: \mathbb{R}^{L \times C} \rightarrow \mathbb{R}^{K}$ mapping inputs to $K$ classes. Following recent advances in vision transformers~\cite{dosovitskiy2020image} and time series models~\cite{cordonnier2021differentiable}, we adopt a patch-based approach to achieve computational efficiency, reducing attention complexity from $O(L^2)$ to $O(N^2)$ where $N = L/p$ and $p$ is the patch size.

\begin{figure}
  \centering
  \includegraphics[width=\textwidth]{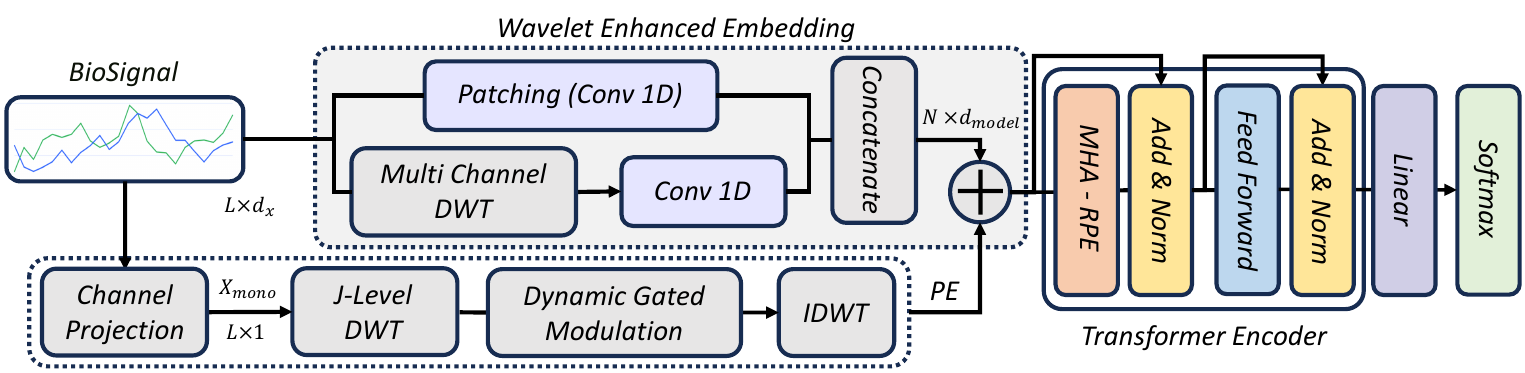}
  \caption{Overall architecture of the WaveFormer model.
}
  \label{fig:arch}
\end{figure}

We propose WaveFormer, a transformer architecture that incorporates wavelet analysis at embedding and positional encoding stages, augmented with relative positional encoding in the attention mechanism. The architecture consists of four main components: wavelet-enhanced patch embedding, signal-aware positional encoding, transformer encoder with relative positional bias, and classification head.
Figure~\ref{fig:arch} illustrates the complete architecture.

\subsection{Wavelet-Enhanced Patch Embedding}

Our embedding layer processes the input signal through two parallel paths that are subsequently fused:

\paragraph{Path A: Patches directly from Raw Signal.}
We apply standard convolutional patching to the original signal $\mathbf{X}$. After padding (if necessary) to ensure $L$ is divisible by patch size $p$, we apply a 1D convolution:
producing $\mathbf{R} \in \mathbb{R}^{d/2 \times N}$ where $N = L/p$ is the number of patches and $d$ is the embedding dimension. These raw patches capture local temporal patterns directly from the input.

\paragraph{Path B: Wavelet-Derived Frequency Features.}

We apply the Discrete Wavelet Transform (DWT). For multi-channel signals, we use grouped convolution to apply DWT independently to each channel:

\begin{align}
cA[c, n] &= \sum_{k} h[k] \cdot \mathbf{X}^T[c, 2n + k] \\
cD[c, n] &= \sum_{k} g[k] \cdot \mathbf{X}^T[c, 2n + k]
\end{align}

\begin{equation}
\mathbf{W}_{\text{input}}[c, n] = cA[c, n] + \alpha \cdot cD[c, n]
\label{eq:wavelet_combine}
\end{equation}


\textbf{Fusion:} The raw and wavelet-derived features are concatenated along the embedding dimension:
\begin{equation}
\mathbf{E}_{\text{patches}} = [\text{Conv1d}(\mathbf{X}^T); \text{Conv1d}(\mathbf{W}_{\text{input}})]^T \in \mathbb{R}^{N \times d}
\label{eq:patch_concat}
\end{equation}

producing $N$ token embeddings, each of dimension $d$.

\subsection{Signal-Aware Positional Encoding}

Standard transformers use signal-agnostic positional encodings that assign identical position information regardless of local temporal characteristics. WaveFormer instead employs Dynamic Wavelet Positional Encoding (DyWPE), which generates position embeddings directly from the signal's frequency content. 

We first apply a learnable channel projection to aggregate information across channels:
\begin{equation}
x_{mono} = x \cdot w_{channel}
\end{equation}
where $w_{channel} \in {R}^{d_x}$ is a learnable projection vector capturing the most relevant temporal dynamics across input channels.

We then apply $J$-level Discrete Wavelet Transform (DWT) to decompose the signal into approximation coefficients $cA_J$ and detail coefficients $\{cD_j\}_{j=1}^J$, capturing frequency content at multiple resolutions:
\begin{equation}
(cA_J, [cD_J, cD_{J-1}, ..., cD_1]) = \text{DWT}(x_{mono})
\end{equation}
The core innovation is the dynamic modulation mechanism:
\begin{equation}
\text{gate}(e, c) = \left( \sigma(W_g e) \odot \tanh(W_v e) \right) \otimes c'
\end{equation}
This generates modulated coefficients for all scales:
\begin{equation}
Yl_{\text{mod}} = \text{gate}(e_{A_J}, cA_J) 
\end{equation}
\begin{equation}
Yh_{\text{mod}} = [\text{gate}(e_{D_J}, cD_J), ..., \text{gate}(e_{D_1}, cD_1)]
\end{equation}
The final positional encoding is reconstructed using Inverse DWT:
\begin{equation}
P_{DyWPE} = \text{IDWT}(Yl_{mod}, Yh_{mod})
\end{equation}
The complete embedding combines token content with signal-aware position information:
\begin{equation}
\mathbf{E}_{\text{final},i} = \mathbf{E}_{\text{input},i} + \mathbf{P}_{\text{DyWPE},i}
\end{equation}



\subsection{Transformer Encoder with Relative Positional Biases}
\label{RPE}

The embeddings output $\mathbf{E}_{\text{final}} \in \mathbb{R}^{(N+1) \times d}$ are processed by $\mathcal{L}$ standard Transformer encoder layers \cite{vaswani2017attention}. Each layer $\ell$ consists of multi-head self-attention (MHSA) and a feed-forward network (FFN), with residual connections and layer normalization:

While DyWPE provides signal-aware positioning, relative temporal relationships are equally important for capturing local dependencies in time series. We augment the multi-head attention mechanism with relative positional encoding (RPE) by introducing learned biases into attention scores:

\begin{equation}
\text{Attention}(\mathbf{Q}, \mathbf{K}, \mathbf{V}) = 
\text{Softmax}\left(\frac{\mathbf{Q}\mathbf{K}^T}{\sqrt{d_k}} + 
\mathbf{B}_{\text{rel}}\right) \mathbf{V}
\end{equation}

where $\mathbf{B}_{\text{rel}}[i,j]$ is a learned bias based on the relative temporal distance between tokens $i$ and $j$.

\begin{equation}
\text{bucket}(r) = \begin{cases}
r & \text{if } |r| < r_{\text{max}}/2 \\
r_{\text{max}}/2 + \lfloor \log_2(|r|/(r_{\text{max}}/2)) \cdot (B/2 - r_{\text{max}}/2) \rfloor & \text{otherwise}
\end{cases}
\end{equation}
where $B$ is the total number of buckets (typically 32) and $r_{\text{max}}$ is the maximum exact distance (typically 16). Each bucket has a learnable bias $b_{ij} \in \mathbb{R}$ shared across all positions with the same relative distance.

After $\mathcal{L}$ layers, we extract the class token representation $\mathbf{h}_{\text{CLS}} = \mathbf{E}^{(\mathcal{L})}[0]$ and apply a two-layer classification head:
\begin{equation}
\mathbf{z} = \mathbf{W}_2 \cdot \text{Dropout}(\text{GELU}(\mathbf{W}_1 \mathbf{h}_{\text{CLS}}))
\end{equation}
where $\mathbf{W}_1 \in \mathbb{R}^{d \times 4d}$, $\mathbf{W}_2 \in \mathbb{R}^{4d \times K}$. Training minimizes cross-entropy loss using Adam optimizer, cosine annealing, gradient clipping, and early stopping.





\section{Experimental Results}

In this section, we evaluate the performance of our model on eight multivariate time series datasets from the UEA archive~\cite{bagnall2018uea} and compare it with the state-of-the-art models. We conducted all experiments on a high-performance Linux server with the following specifications: an AMD EPYC 7513 32-Core Processor (2 sockets, 128 threads), one NVIDIA RTX A5000 GPUs (24GB GDDR6 each, CUDA version 12.2, Driver version 535.113.01), 503GB of RAM, and a dual storage system consisting of an 892.7GB NVMe SSD and a 10.5TB HDD. The operating system used was Ubuntu 20.04.6 LTS (Kernel 5.4.0-200-generic). 
All experiments use consistent hyperparameters: 4 transformer layers, 4 attention heads, 128 hidden dimensions, dropout rate 0.2, and an Adam optimizer. The complete code implementation and benchmarks are made publicly available for reproducibility: {\small \url{https://github.com/imics-lab/waveformer}}.

\subsection{Datasets}

Two Human Activity Recognition, five brain signal and Heartbeat signal datasets have been used in these experiments, as shown in table \ref{tab:datasets}. Datasets span 50–3000 timesteps, 1–144 channels, and 200–7352 samples, providing diverse temporal and spectral characteristics for comprehensive evaluation.





\begin{table}

\caption{Time series dataset properties.}
\vspace{1mm}
\label{tab:datasets}
\fontsize{8.5}{10}\selectfont
\setlength{\tabcolsep}{5.5pt}
\renewcommand{\arraystretch}{1.1}
\begin{tabular}{@{}l c c c c c c@{}}
\toprule
\textbf{Dataset} & \textbf{Train} & \textbf{Test} & \textbf{Length} & \textbf{Class} & \textbf{Channel} & \textbf{Type} \\
\midrule
WalkingSittingStanding (WSS) & 7352 & 2947 & 206 & 6 & 3 & HAR\\
UWaveGestureLibraryAll (UWG) & 896 & 3582 & 945 & 8 & 1 & HAR\\
FaceDetection (FD) & 5890 & 3524 & 62 & 2 & 144 & EEG \\
FingerMovements (FM) & 316 & 100 & 50 & 2 & 28 & EEG \\
SelfRegulationSCP1 (SR1) & 268 & 293 & 896 & 2 & 6 & EEG \\
SelfRegulationSCP2 (SR2) & 200 & 180 & 1152 & 2 & 6 & EEG \\
MotorImagery (MI) & 278 & 100 & 3000 & 2 & 64 & EEG \\
Heartbeat  & 204 & 205 & 405 & 2 & 61 & AUDIO \\

\bottomrule
\end{tabular}
\begin{flushleft}
\footnotesize
\end{flushleft}

\end{table}

\subsection{Comparing with state-of-the-art models}
We compare WaveFormer against five established state-of-the-art time series models representing distinct architectural paradigms: 
\begin{itemize}
    \item \textbf{ResNet}\cite{wang2017time}, a deep residual CNN baseline.
    \item \textbf{InceptionTime}\cite{fawaz2020inceptiontime}, an ensemble of multi-scale inception modules. 
    \item \textbf{TST}\cite{zerveas2021transformer}, a vanilla transformer for time series.
    \item \textbf{PatchTST}\cite{cordonnier2021differentiable}, a patch-based transformer baseline.
    \item \textbf{ConvTran}\cite{foumani2024improving}, a hybrid transformer model with absolute and relative positional encodings.
\end{itemize}

These baselines collectively span CNN-based, transformer-based and hybrid models, enabling a comprehensive assessment of WaveFormer's contributions. 


Table~\ref{tab:results} presents classification accuracy across all datasets and models. WaveFormer achieves the best performance on seven out of eight datasets, with particularly strong results on small-sample scenarios: SelfRegulationSCP2 (200 samples, 61.1\% accuracy), MotorImagery (278 samples, 64.0\% accuracy), and Heartbeat (204 samples, 78.4\% accuracy). On larger datasets, WaveFormer attains 91.3\% accuracy on WalkingSittingStanding (7,352 samples) and 93.0\% on UWaveGestureLibrary (896 samples), demonstrating effectiveness across different data regimes.

\begin{table}
\caption{Average accuracy of six deep learning based models over eight multivariate time series datasets.}
\vspace{1mm}
\label{tab:results}
\fontsize{8}{8}\selectfont
\setlength{\tabcolsep}{6.2pt}
\renewcommand{\arraystretch}{1.1}
\begin{tabular}{@{}l  c c c c c c@{}}
\toprule
\textbf{Data} & \textbf{WaveFormer} & \textbf{ConvTran} & \textbf{PatchTST} & \textbf{IT} & \textbf{TST} & \textbf{ResNet} \\
\toprule
WalkingSittingStanding  & \textbf{0.913} & 0.905 & 0.897  & 0.873 & 0.892 & 0.871 \\
FaceDetection & 0.658 & \textbf{0.663} & 0.654  & 0.575 & 0.647 & 0.584 \\
UWaveGestureLibraryAll & \textbf{0.930} & 0.899 & 0.879  & 0.906 & 0.886 & 0.847 \\
FingerMovements & \textbf{0.580} & 0.560 & \textbf{0.580}  & 0.560 & 0.560 & 0.540 \\
MotorImagery & \textbf{0.640} & 0.560 & 0.530  & 0.530 & 0.480 & 0.520 \\
SelfRegulationSCP1 & \textbf{0.918} & \textbf{0.918} & 0.872  & 0.866 & 0.864 & 0.843 \\
Heartbeat & \textbf{0.784} & 0.778 & 0.705  & 0.629 & 0.708 & 0.724 \\
SelfRegulationSCP2 & \textbf{0.611} & 0.591 & 0.551  & 0.482 & 0.533 & 0.512 \\
\bottomrule
\end{tabular}
\begin{flushleft}
\footnotesize
\end{flushleft}
\end{table}


\subsection{Ablation Study}

In this section, to validate each architectural component's contribution, we conduct a comprehensive ablation experiment on four representative datasets spanning different signal modalities: WSS (HAR), UWaveGesture (HAR), SelfRegulationSCP1 (EEG), and Heartbeat (Audio). We systematically remove individual components and measure the impact on classification accuracy.

The full WaveFormer model achieves 88.7\% mean accuracy across the four datasets. When the wavelet-enhanced embedding is removed (using only raw time-domain patches), performance drops by 3.1 percentage points to 85.6\%, representing the largest individual contribution. This demonstrates that explicit frequency-domain decomposition via DWT provides complementary information that standard convolutional patching cannot capture. Removing DyWPE while retaining standard learnable positional encoding results in a 2.6pp decrease (86.1\%), indicating that signal-adaptive position modulation improves upon fixed positional representations. Finally, removing RPE causes a 1.8pp drop (86.9\%), showing that relative positional bias, while having the smallest individual effect, still contributes meaningfully to overall performance.

Then we compare the different types of attention mechanism with relative positional encoding to choose the best combination with wavelet embedding for WaveFormer in terms of both accuracy and training time.

For this ablation study we run experiment on different type of attention mechanism on WaveFormer architecture.

\begin{figure}
  \centering
  \includegraphics[width=\textwidth]{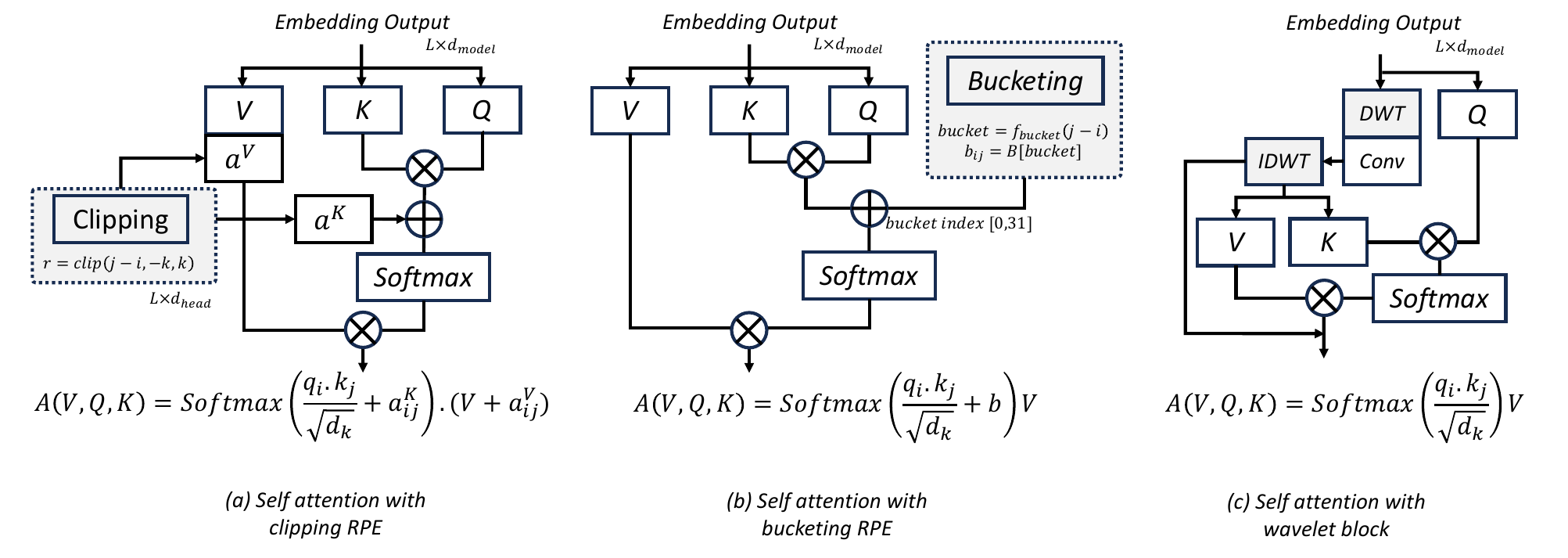}
  \caption{Three approaches to compute self-attention.
         (a) Clipping RPE: relative distances beyond a threshold are clipped.
         (b) T5-based bucketing RPE (adopted in WaveFormer): uses logarithmic 
             bucketing for parameter-efficient distance modeling.
         (c) Wavelet-based attention downsampling.
          }
  \label{fig:rpe_variants}
\end{figure}

Figure~\ref{fig:rpe_variants} illustrates three approaches to computing self-attention. Variant (a) uses simple clipping, where relative distances beyond a threshold are treated identically, this approach is parameter-efficient but loses information about large temporal gaps. Variant (b) employs T5-style bucketing~\cite{raffel2020exploring}: short lags use linear bucketing for fine-grained resolution (e.g., distances 1, 2, 3 each get separate parameters), while longer distances use logarithmic bucketing for parameter efficiency (e.g., distances 32-63, 64-127 share parameters). This provides a favorable trade-off between expressiveness and parameter count. Variant (c) shows an alternative where attention keys and values are downsampled via wavelet decomposition before computing attention \cite{yao2022wavevit}. 

As mentioned in section \ref{RPE}, we adopt T5-style bucketing RPE (Variant b) in WaveFormer as it provides the best accuracy-efficiency trade-off. Compared to clipping (Variant a), bucketing maintains expressiveness for long sequences without parameter explosion. Compared to wavelet-downsampled attention (Variant c), bucketing preserves full-resolution features while still achieving $\mathcal{O}(N \log L)$ computational complexity through efficient bucketing. Importantly, this choice is orthogonal to our wavelet-enhanced embedding: we apply wavelets at the input stage to explicitly capture frequency information, while using bucketing RPE for efficient relative position modeling within the transformer.

\subsection{Performance Analysis}

\begin{figure}
   \centering
   \begin{subfigure}[b]{0.47\columnwidth}
       \centering
       \includegraphics[width=\textwidth]{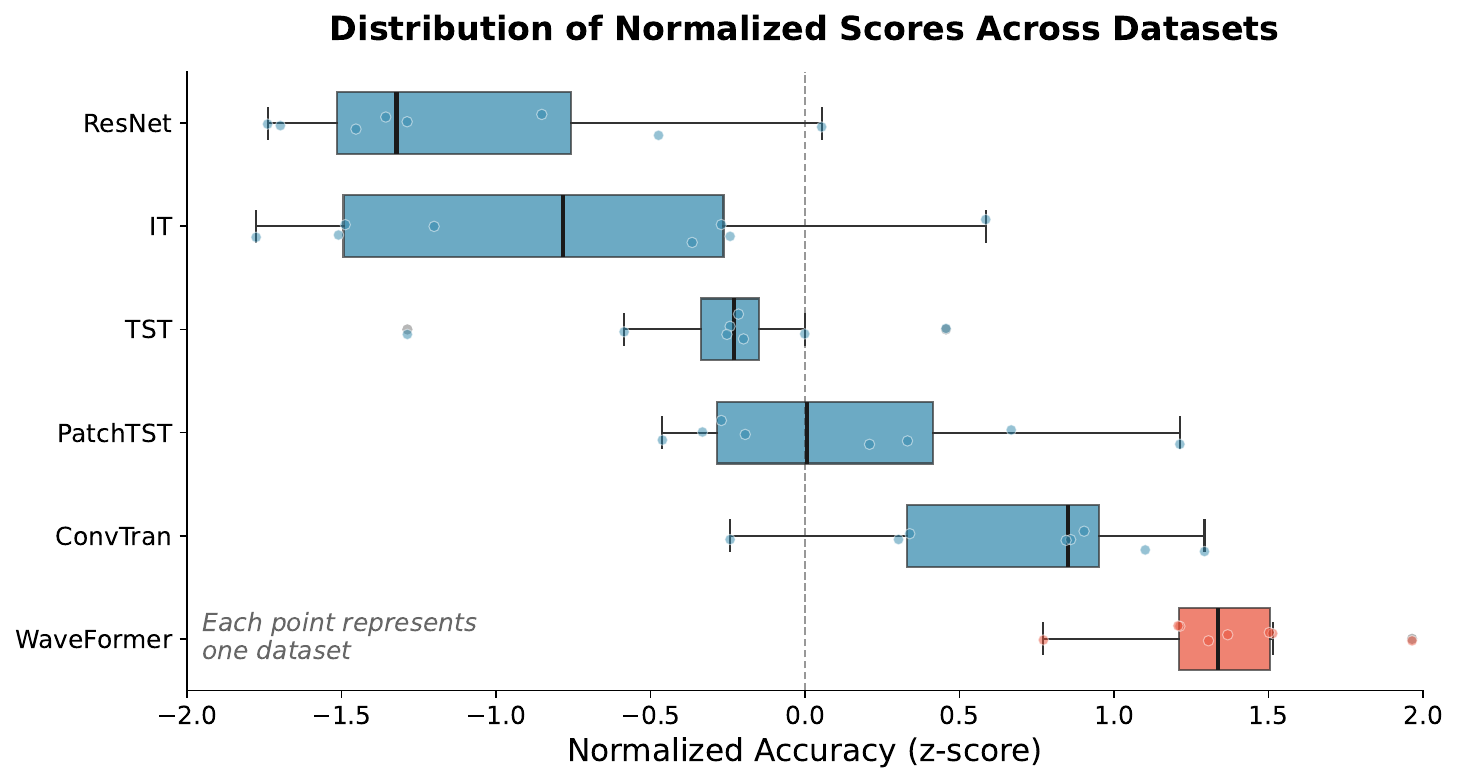}
       \caption{\small Distribution of z-score normalized (mean = 0, std = 1) classification accuracy across 8 datasets for each model. Box plots show the median (dark line), interquartile range (box), and data range (whiskers), with outliers shown as isolated points.}
       \label{fig:box}
   \end{subfigure}
   \hfill
   \begin{subfigure}[b]{0.48\columnwidth}
       \centering
       \includegraphics[width=\textwidth]{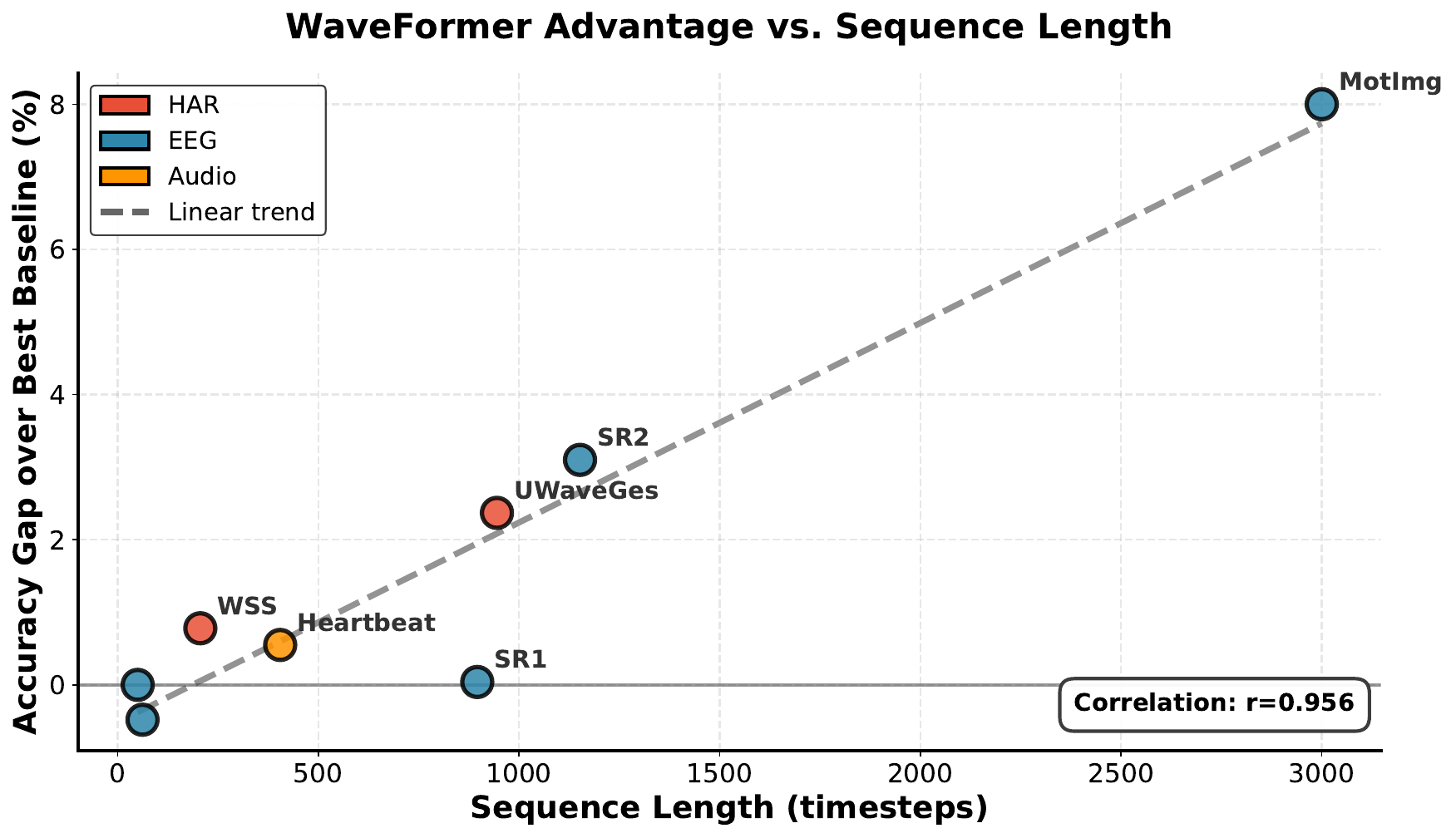}
       \caption{\small Performance advantage versus sequence length. WaveFormer's accuracy improvement over the best baseline correlates positively with sequence length (r=0.741), with largest gains on long sequences. Points are colored by signal type: HAR (red), EEG (blue), Audio (orange).}
       \label{fig:length}
   \end{subfigure}
   \caption{\small WaveFormer performance analysis.}
   \label{fig:performance}
\end{figure}

Figure~\ref{fig:performance} presents a comprehensive analysis of WaveFormer's performance characteristics. The normalized accuracy distribution (Figure~\ref{fig:box}) demonstrates WaveFormer's consistent superiority across all eight datasets, achieving a median z-score of 1.34 compared to the next best baseline (ConvTran: 0.85). The tight distribution indicates robust performance across diverse signal types and dataset characteristics, with WaveFormer never ranking worse than second place on any dataset. Furthermore, the relationship between sequence length and performance advantage (Figure~\ref{fig:length}) reveals a positive correlation (r=0.741, p=0.035), with WaveFormer's improvements ranging from minimal gains on short sequences (<500 timesteps) to substantial improvements on long sequences ($>1000$ timesteps). This pattern validates our hypothesis that wavelet decomposition's multi-scale analysis becomes increasingly valuable as sequences contain richer temporal structure, with the largest gain observed on MotorImagery (3000 timesteps, +8.0 percentage points).


Signal complexity (channel count) shows moderate correlation: complex signals ($>10$ channels) improve +2.02pp versus +1.57pp for simpler signals. Multi-channel DWT via grouped convolution preserves spatial information while extracting per-channel frequency content. However, sequence length dominates: single-channel UWaveGes (945 steps) gains +2.37pp from periodic motion patterns at multiple timescales, while high-dimensional FaceDetection (144 channels, 62 steps) shows -0.48pp due to insufficient sequence length.



\section{Conclusion}

In this paper we presented WaveFormer, a transformer that integrates wavelet decomposition at two stages for comprehensive multi-resolution time series analysis. 
Evaluation on eight datasets spanning human activity recognition and EEG signals demonstrates competitive performance. The architecture successfully handles diverse temporal and spectral characteristics, from short/wide sequences (FingerMovements) to long sequences (MotorImagery).

Our work establishes that explicit frequency-domain processing through wavelet transforms, when properly integrated into transformers, provides an effective approach for time series classification. The dual-stage design offers a generalizable framework applicable to any domain where multi-scale patterns are relevant.


\balance

\bibliographystyle{splncs04}
\bibliography{references}

\end{document}